# Power, Control, and Data Acquisition Systems for Rectal Simulator Integrated with Soft Pouch Actuators

Zebing Mao[1*], Sota Suzuki[2], Ardi Wiranata[3], Junji Ohgi [1], Shoko Miyagawa[4],

[1]Faculty of Engineering, Yamaguchi University, Yamaguchi, Japan
[2]School of Engineering, Tokyo Institute of Technology, Tokyo, Japan
[3]Department of Mechanical and Industrial Engineering, Universitas Gadjah Mada, Yogyakarta, Indonesia
[4]Faculty of Nursing and Medical Care, Keio University, Kanagawa, Japan

[*]mao.z.aa@yamguchi-u.ac.jp

## Abstract

Fecal incontinence (FI) is a significant health issue with various underlying causes. Research in this field is limited by social stigma and the lack of effective replication models. To address these challenges, we developed a sophisticated rectal simulator that integrates power, control, and data acquisition systems with soft pouch actuators. The system comprises four key subsystems: mechanical, electrical, pneumatic, and control and data acquisition. The mechanical subsystem utilizes common materials such as aluminum frames, wooden boards, and compact structural components to facilitate the installation and adjustment of electrical and control components. The electrical subsystem supplies power to regulators and sensors. The pneumatic system provides compressed air to actuators, enabling the simulation of FI. The control and data acquisition subsystem collects pressure data and regulates actuator movement. This comprehensive approach allows the robot to accurately replicate human defecation, managing various feces types including liquid, solid, and extremely solid. This innovation enhances our understanding of defecation and holds potential for advancing quality-of-life devices related to this condition.

## Specifications table

| | |
|---|---|
| Hardware name | Rectal biomedical simulator |
| Subject area | Engineering and materials science |
| Hardware type | Measuring physical properties and in-lab sensors |
| Closest commercial analog | Rectal simulator mimicking the human defecation process |
| Open-source license | CC-By Attribution 4.0 International |
| Cost of hardware | $ 4931.58 |
| Source file repository | https://osf.io/yv83a/files/osfstorage |

## 1. Hardware in context

Recently, physiological organ simulators have provided a novel approach to assist and accelerate the evaluation of numerous therapeutic concepts in the early development stages, as well as to deepen the understanding of the biomechanics associated with various diseases [1][2][3][4][5][6]. In simulators, soft robotics offer a new method for mimicking or assisting the functions of human organs. For example, implantable cardiac soft robotic devices can enhance cardiac function in patients with isolated left or right heart failure [7]; gastric robots can serve as testing environments for designing and evaluating innovative food products and pharmaceuticals [8]. Although soft robotics in simulated organs have demonstrated their value as supportive and complementary methods for mathematical modeling, exploring human physiology, and validating medical procedures, there is limited focus on replicating the defecation system due to stigma and social taboos. Consequently, there is significant interest in developing physical models that replicate the defecation system and create artificial anal sphincter (AAS) devices [9].

Researchers have explored various actuation technologies, such as rigid clamping mechanisms, fluid actuators, cable-driven actuators, magnetic actuators, and shape memory alloy actuators, to investigate their ability to reproduce the sphincter pressures observed clinically [10][11][12][13][14]. Additionally, other actuators have shown potential for occluding the anal canal [15][16]. Although current research primarily focuses on AAS devices, there is still limited research dedicated to replicating the defecation system and investigating mechanisms related to fecal incontinence (FI). William E. Stokes and colleagues proposed a method for manufacturing, measuring, and controlling a physical simulator of the human defecation system to study the individual and combined effects of anal-rectal angles and sphincter pressures on defecation continuity [17]. This study utilized rigid components such as stepper motors and linear platforms as actuators but failed to replicate the peristaltic movement of the rectum and the opening-closing movement of the anus. Koushi Tokoro and others developed a robotic defecation simulator (defecation robot) that can simulate defecation involving the rectum, anal sphincter, puborectalis muscle, and abdominal pressure [18]. Therefore, we propose a power, control, and data acquisition system for a rectal simulator integrated with soft bag actuators, which can replicate abdominal pressure, rhythmic peristalsis, and the opening-closing movements of the anus on a symmetrical level [19] (**Fig. 1**). **Fig. 1**a shows the entire setup, which includes a rectal simulator mounted on a stand and an air compressor. In **Fig. 1**b, we can find that individual main components in our device: rectum mold, flush and stand. **Fig. 1**c focuses on the pneumatic, control and monitoring systems. Inside the frame, it contains an Arduino microcontroller, a voltage supply, pressure sensors, release valves, regulator, filter, and connections to LabVIEW software for data acquisition and control.

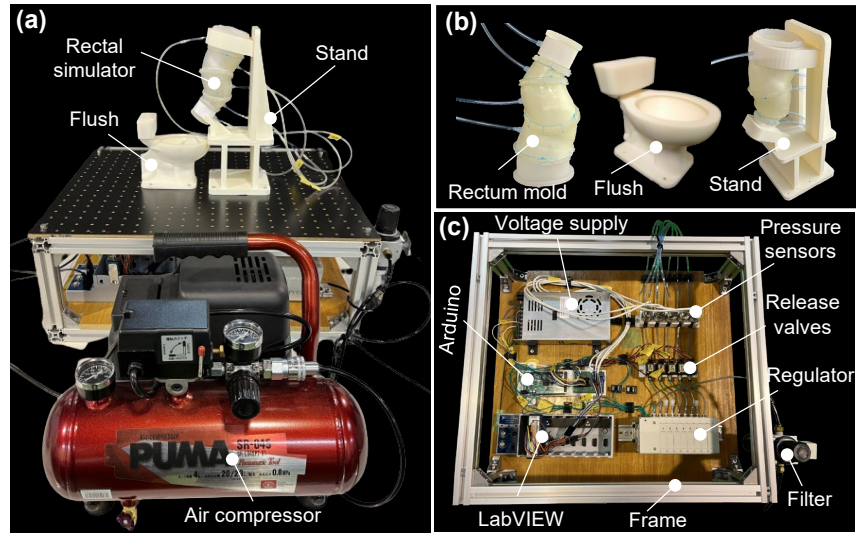

**Fig. 1** Optical images of the rectal simulator integrated with soft pouch actuators and its system. **(a)** Hardware components of the rectal simulator. **(b)** Rectum mold, flush mechanism, and stand. **(c)** Electrical, pneumatic, and control hardware.

## 2. Hardware description

This section presents the detailed mechanical and electrical design of an advanced electromechanical system. Utilizing a combination of large aluminum frames and wooden components, we constructed a robust rectangular framework to securely house various system elements. Key components, including the DC power supply, solenoid valves, and sensors, were mounted on custom-designed bases created through 3D printing technology. The system integrates an air compressor, pressure sensors, and an Arduino-based control module to manage pneumatic and electrical functions. A graphical user interface (GUI) for data acquisition (DAQ) enhances real-time monitoring and control.

### 2.1 Mechanical design

To facilitate the installation of various components of this electromechanical system, we employed large-sized aluminum frames and smaller components like wooden boards to construct a rectangular box-like framework. This framework primarily consists of four short aluminum frames, each measuring 150 mm, and eight long aluminum frames, each measuring 450 mm. Additionally, we used several D brackets, T-nuts, and hexagon socket bolts to connect the short and long aluminum frames. The ordinary plywood in **Fig. 2**a features a few holes, allowing for the later attachment of components such as the DC power supply, solenoid valves, and switch valves. In contrast, the perforated board has numerous through-holes to facilitate the attachment of the flush and stand using bolts and nuts. Therefore, we designed three-dimensional models of the flush and stand, as shown in **Fig. 2**b and **Fig. 2**c. To securely fix the rectal model, we designed Stand 1 to fit with the anus of the rectal model, and Stand 2 to connect with the S-shaped rectal opening. Stand 3 is used to enhance stability. Stand 4 was designed to increase the height of the model. Stand 5 was designed to connect the stands with the perforated board.

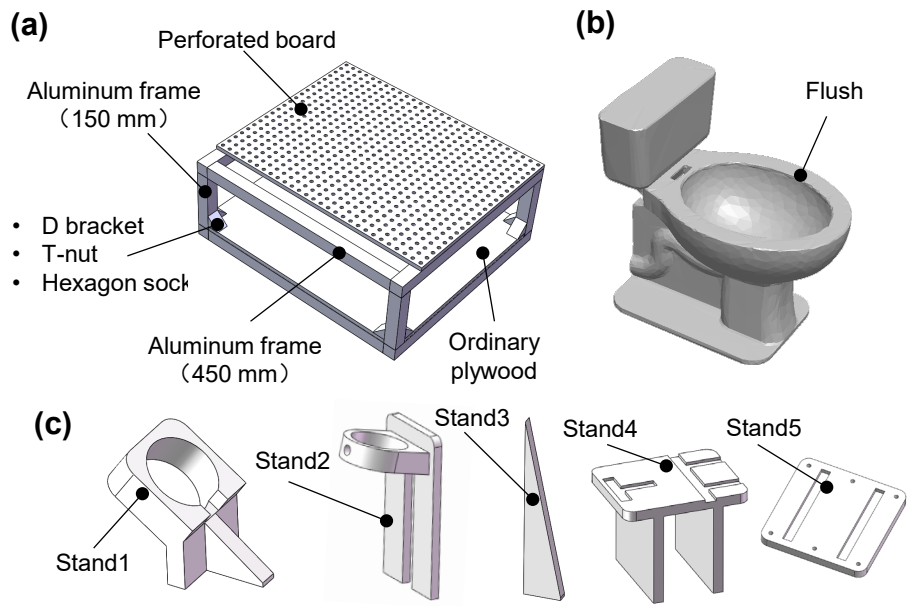

**Fig. 2 (a)** Design of the frame including aluminum frame, board, and ordinary plywood; **(b)** Three-dimensional design of the flush; **(c)** Design of the various parts of the stand.

In addition, we designed bases for various components such as the valve body (**Fig. 3**a), sensors (**Fig. 3**b), DC power supply fixation (**Fig. 3**c), controller and PCB circuit board fixator (**Fig. 3**d). Each component's base is crucial to ensure the stability, functionality, and overall efficiency of the entire system. The fixation for the DC power supply was another critical component. A stable and secure base for the DC power supply is essential to prevent any electrical disturbances or interruptions in power delivery. Similarly, the bases for the sensors were designed to provide a secure and stable platform. Sensors play a pivotal role in monitoring various parameters, and their accurate readings are vital for the system's performance. Also, the controller, often an Arduino in our setup, and the PCB board, which hosts various electronic circuits, require a stable and well-organized platform. These models were then translated into physical bases using 3D printing technologies, ensuring high accuracy and repeatability in the fabrication process.

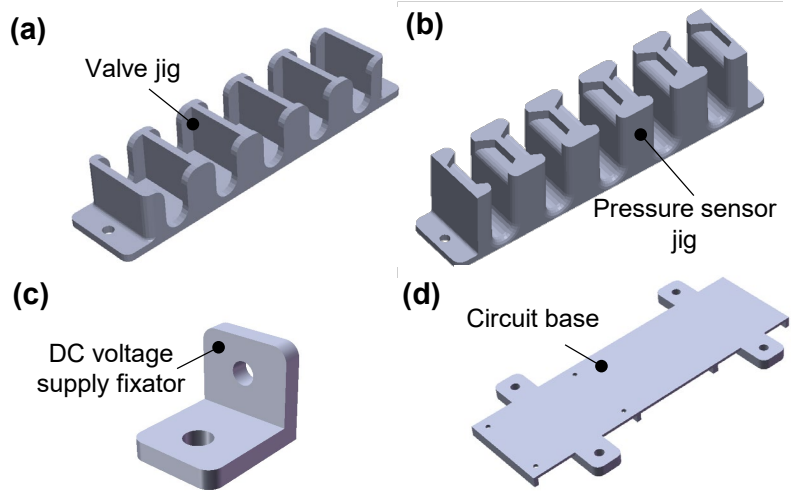

**Fig. 3 (a)** Base for fixing the valve; **(b)** Base for fixing the pressure sensor; **(c)** DC voltage supply fixation; **(d)** Base for fixing the controller Arduino and PCB board.

## 2.2 Power electrical, control and data acquisition systems

The systems can be categorized into several types: compressor and pneumatic components; electrical and electronic components; pressure sensors and power supply; control and interface modules; fasteners and connectors; and miscellaneous electronic components (**Fig. 4**). The air compressor, a core part of our system, provides the necessary compressed air to drive the entire pneumatic system (**Fig. 4**a). Complementing the compressor is the air filter regulator, which filters and regulates the air pressure, ensuring that the system operates within the desired parameters (**Fig. 4**b). **Fig. 4**c and **Fig. 4**d showcase electro-pneumatic regulator and solenoid valve, respectively. The electro-pneumatic regulator allows for precise control while the solenoid valve helps discharge the air quickly. Also, the perfboard is used for mounting electronic components like transistors and electrical connectors. Considering the pressure sensors and power supply (**Fig. 4**e & **Fig. 4**f), The pressure sensors used to pressure in the pneumatic system, providing critical data for system operation. The power supply unit used to provide a stable 24V DC voltage to pressure sensors, regulator and other electronic components. Besides, we used Arduino microcontroller to control pressure systems and the switching ability of the solenoid valve (**Fig. 4**g). The DAQ (Data acquisition) modules interface with the sensors and actuators, facilitating data collection and control signal distribution (**Fig. 4**h). In regard with the fasteners and connectors, bolts and screws are used to assemble various mechanical parts of the system like aluminum frame (**Fig. 4**i). Electrical wires and cables are used for connecting electronic components (**Fig. 4**j). Pneumatic connectors used for joining pneumatic tubes and components (**Fig. 4**k&**Fig. 4**l). The Integrated Circuit (IC) chips are used to control the ON-OFF functions of the solenoid valve via the controlling process of Arduino microcontroller.

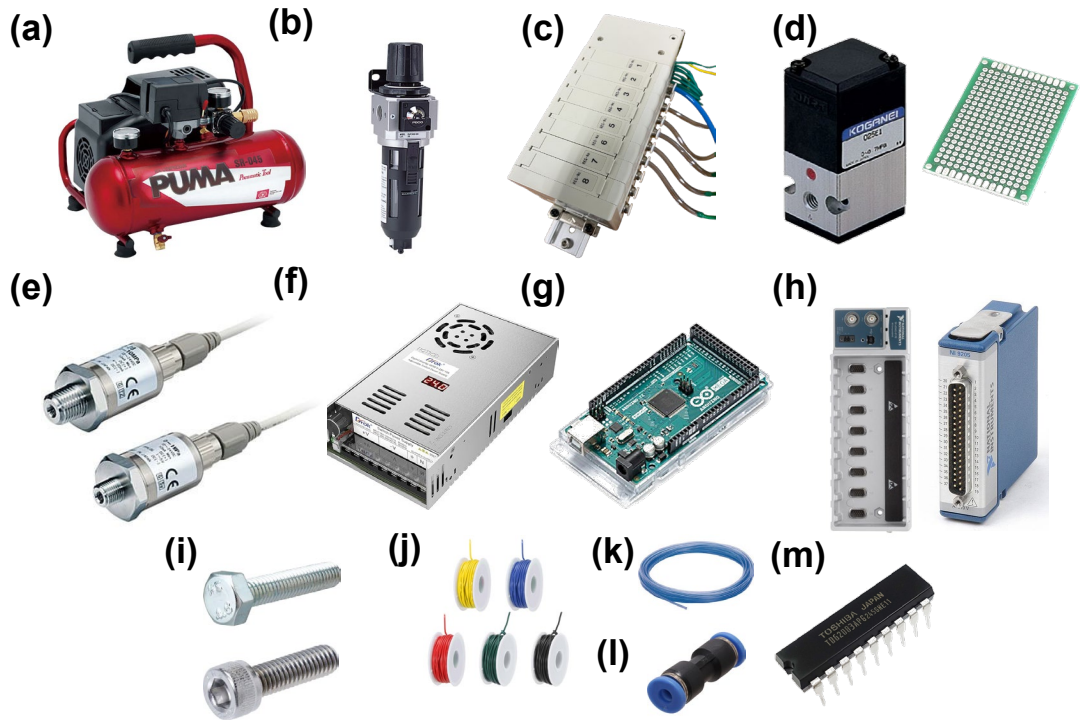

**Fig. 4** Components of electrical, control and data acquisition systems. **(a)** Air compressor; **(b)** Air filter regulator; **(c)** Electro-pneumatic regulator; **(d)** Solenoid valve and perfboard; **(e)** Pressure sensors; **(f)** DC power supply; **(g)** Arduino microcontroller; **(h)** DAQ Modules; **(i)** Bolts; **(j)** Wires and cables; **(k)** Tubes; **(l)** Adaptors; **(m)** Transistor array.

Fig. 5 illustrates the connection layout for a multi-system setup including the power system, pneumatic circuit, and the electric, control, and data acquisition (DAQ) systems. Each subsystem is integrated to ensure efficient functionality and communication within the entire setup. The power system includes a 110V 50Hz power supply connected to various components such as LabVIEW, Arduino, DC power supply, and an air compressor (Fig. 5a). The DC Power Supply provides necessary power to the Arduino and the transistor Array. The pneumatic circuit comprises an air compressor connected to a regulator which controls the air pressure. This regulated air is then distributed to various pressure sensors and release valves. The pressure sensors (P1 to P5) monitor the air pressure in different sections of the pneumatic circuit, and release valves (R1 to R5) control the airflow to respective actuators (A1 to A5) (Fig. 5b). The electric and control system centers around the Arduino, which interfaces with the Transistor Array and the release valves (Fig. 5c). The Transistor Array controls the opening and closing of the release valves (R1 to R5) based on signals from the Arduino (I1 to I5). The sensors (P1 to P5) feed data back to the NI9205 DAQ system, which communicates with LabVIEW for real-time monitoring and control. Ground connections (GND) ensure a common reference point for all components, ensuring stable operation.

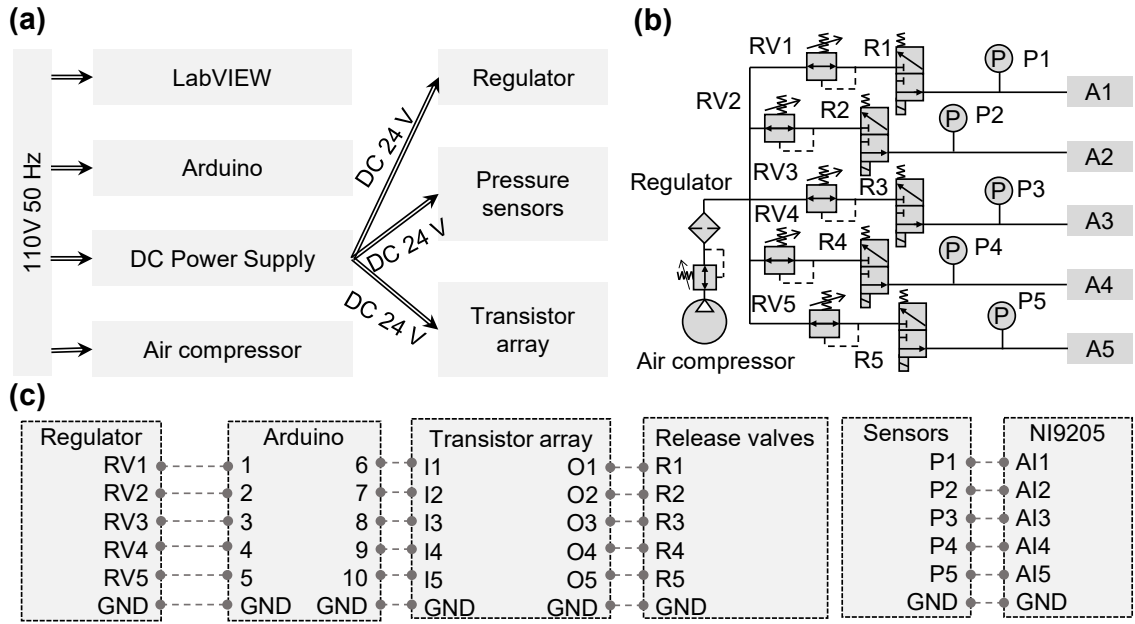

**Fig. 5** Connection for each system. **(a)** Power system; **(b)** Pneumatic circuit; **(c)** Electric, control system and DAQ system

Fig. 6 showcases the graphical user interface (GUI) designed for data acquisition (DAQ) systems. This interface is crucial for real-time monitoring and analysis of sensor data, providing a comprehensive and user-friendly platform for interaction with the DAQ system. Key components of the interface included several parts. (a) Path and Configuration Settings. At the top left, users can specify the file path for saving the data (e.g., "C:\Users\havi\Desktop\Y6.csv"). This ensures that data is properly logged for further analysis. The sample rate setting, currently at 1000, allows users to control the frequency at which data is sampled. Additionally, the terminal configuration can be adjusted, with "default" being the current setting. (b) Channel selection. The GUI provides input fields (P1 to P5) for each channel, allowing users to designate specific DAQ channels (e.g., "dDAQ2Mod2/ai1") to monitor and record data from different sensors. This flexibility enables the monitoring of multiple data streams simultaneously. (c) Real-time data visualization. The main part of the GUI features five graphical plots, each corresponding to a different channel (P1 to P5). These plots display the amplitude of the signal over time, providing a visual representation of sensor data. (d) Control buttons. At the bottom left, a prominent "Stop" button is available for users to halt data acquisition immediately. (e) Numeric Displays. In the bottom right section, numeric displays provide real-time values for each channel (P1 to P5). These displays offer quick reference points for users to monitor sensor outputs without solely relying on the graphical plots. Overall, this GUI is an integral component of the DAQ system, facilitating effective data management, visualization, and control. It enhances the usability of the system by providing intuitive controls and real-time feedback, crucial for conducting precise and reliable data acquisition in various experimental settings.

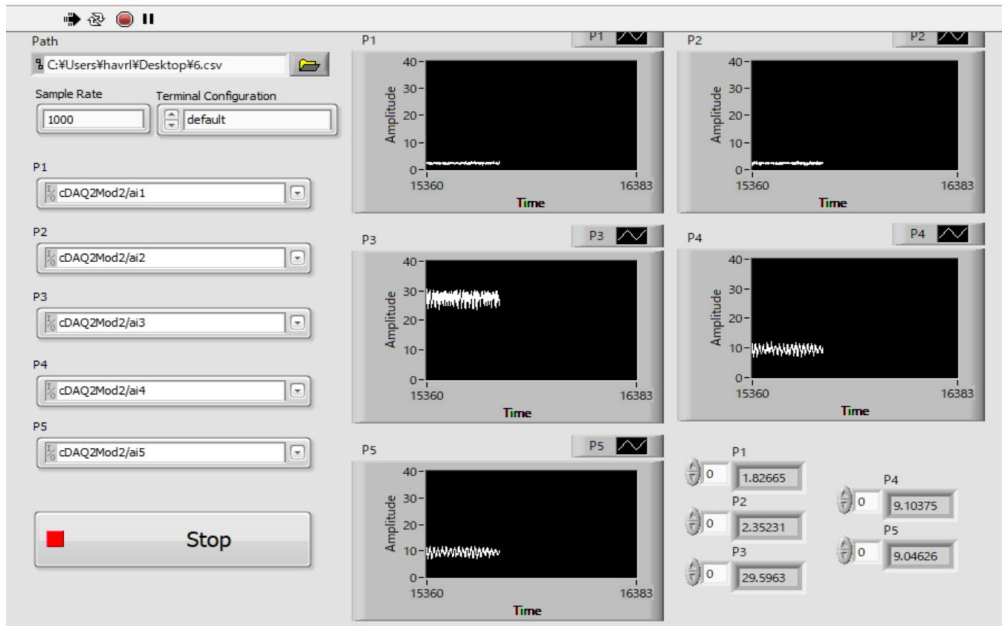

**Fig. 6** Graphic user interface (GUI) for DAQ system

## 3. Design files summary

Table 1
Design file name and summary

| Design file name | File type | Open source license | Location of the file |
|---|---|---|---|
| 1-stage (**Fig. 2**a) | CAD files | CC BY-4.0 | https://osf.io/srf2g |
| 2-flush (**Fig. 2**b) | CAD files | CC BY-4.0 | https://osf.io/c4nrg |
| Stand1 (**Fig. 2**c) | CAD files | CC BY-4.0 | https://osf.io/ydm7j |
| Stand2 (**Fig. 2**c) | CAD files | CC BY-4.0 | https://osf.io/7hkyr |
| Stand3 (**Fig. 2**c) | CAD files | CC BY-4.0 | https://osf.io/rd69c |
| Stand4 (**Fig. 2**c) | CAD files | CC BY-4.0 | https://osf.io/xstc9 |
| Stand5 (**Fig. 2**c) | CAD files | CC BY-4.0 | https://osf.io/qnysk |
| 1-valve jig (**Fig. 3**a) | CAD files | CC BY-4.0 | https://osf.io/ntsbj |
| 2-Pressure sensor jig (**Fig. 3**b) | CAD files | CC BY-4.0 | https://osf.io/93dxg |
| 3-DC voltage supply fixation (**Fig. 3**c) | CAD files | CC BY-4.0 | https://osf.io/39v42 |
| 4-electronic base (**Fig. 3**d) | CAD files | CC BY-4.0 | https://osf.io/5kqa7 |

| | | | | |
|---|---|---|---|---|
| LabVIEW code | Vi files | CC BY-4.0 | | https://osf.io/mkj97 |
| Arduino code | Ino files | CC BY-4.0 | | https://osf.io/4a7pt |

## 4. Bill of materials summary

**Table 2**
Material and part list of the system.

| Designator | Component | Number | Cost per unit (USD) | Total cost (USD) | Source of materials |
|---|---|---|---|---|---|
| Aluminum frame (450 mm) (**Fig. 2**a) | SF-30・30 | 8 | 3.45 | 27.6 | https://www.monotaro.com/p/6955/1695/ |
| Aluminum frame (150 mm) (**Fig. 2**a) | SF-20・20 | 4 | 1.20 | 4.80 | https://www.monotaro.com/p/6955/1056/ |
| D bracket (**Fig. 2**a) | SFJ-018 | 24 | 0.76 | 18.24 | https://www.monotaro.com/p/1958/1415/ |
| T-nut (**Fig. 2**a) | SFB-002 | 48 | 0.37 | 17.76 | https://www.monotaro.com/p/1958/1503/ |
| Hexagon socket bolt (**Fig. 2**a) | CS-06-12 | 48 | 0.14 | 6.72 | https://www.monotaro.com/p/0840/6116/ |
| Ordinary plywood (**Fig. 2**a) | T2G2-500-500 | 1 | 8.71 | 8.71 | https://www.monotaro.com/g/04176512/#op_1763=500&op_1764=500 |
| Perforated board (**Fig. 2**a) | 580×445 | 1 | 26.26 | 26.26 | https://item.rakuten.co.jp/mokuzai-o/25uwt4w580d445/ |
| Air compressor (**Fig. 4**a) | SR-L04SPT-01 | 1 | 198.16 | 198.16 | https://www.monotaro.com/p/6684/8223/ |
| Filter regulator (**Fig. 4**b) | FRF300-02-MD | 1 | 54.62 | 54.62 | https://www.pisco.co.jp/cad/l/l09/FRF/ |
| Electro-pneumatic regulator (**Fig. 4**c) | MEVT500-0C6-T11R-8-U-3 | 5 | 302.45 | 1512.25 | https://www.ckd.co.jp/kiki/jp/product/detail/223/MEVT |
| Solenoid valve (**Fig. 4**d) | 025E1-81-PSL | 5 | 27.16 | 135.8 | https://official.koganei.co.jp/ |
| Universal printed circuit board (**Fig. 4**d) | 8x12cm | 1 | 1.25 | 1.25 | https://www.monotaro.com/g/05240665/ |
| Pressure sensors (**Fig. 4**e) | PSE573-1 | 5 | 87.90 | 439.5 | https://www.monotaro.com/p/4238/2926/ |
| DC voltage supply (**Fig. 4**f) | DROK DC 0-24V 0-20A 480W | 1 | 37.1 | 37.1 | https://www.droking.com/ |
| Arduino Mega 2560 (**Fig. 4**g) | ATmega2560 | 1 | 51.17 | 51.17 | https://www.arduino.cc/ |
| CompactDAQ Systems (**Fig. 4**h) | cDAQ-9178 | 1 | 2515 | 2515 | https://www.ni.com/ja.html |
| | NI9205 with DSUB | 1 | 1592 | 1592 | https://www.ni.com/ja.html |
| Hex bolt with full threading (**Fig. 4**i) | M6 ×18 mm | 22 | 0.05 | 1.1 | https://www.monotaro.com/p/0551/7145/?t.q=%83%7B%83%8B%83g |
| Hex socket head cap bolt (**Fig. 4**i) | M6 ×20 mm | 15 | 0.23 | 3.45 | https://www.monotaro.com/p/2901/7976/?t.q=%83%7B%83%8B%83g |
| Electrical wires | PVC-20AWG | 1 | 17.32 | 17.32 | https://www.amazon.co.jp/ |
| Tubes (**Fig. 4**k) | UB0425-20-BU | 1 | 6.96 | 6.96 | https://www.monotaro.com/p/0914/9052/?fem1=1 |
| Adaptors (**Fig. 4**l) | MPUC-4 | 14 | 1.31 | 18.34 | https://www.monotaro.com/p/5288/4458/ |
| Transistor array (**Fig. 4**m) | TD62083APG | 1 | 0.48 | 0.48 | https://akizukidenshi.com/catalog/g/g109634/ |

## 5. Build instructions

For the mechanical structure, we first use bolts, D brackets, and hexagon socket bolts to connect four 450mm long aluminum frames at the base, forming the foundation (**Fig. 2a**). Additionally, we secure ordinary plywood onto the base to mount other components (**Fig. 2a**). We also use bolts and nuts to fix essential components such as the DC power supply (**Fig. 4f**), air filter regulator (**Fig. 4c**), and DAQ modules (cDAQ-9178, **Fig. 4h**) onto the base plate. To enhance the fixation, we use our designed fixtures for components such as the pressure sensor (**Fig. 4e**), valve (**Fig. 4h**), and circuit bases (**Fig. 4d**). Subsequently, we mount the corresponding solenoid valve (**Fig. 4h**), perfboard (**Fig. 4d**), and pressure sensors (**Fig. 4e**) onto our fixtures. Finally, we insert the NI NI9205 module into the cDAQ-9178, completing the installation of the basic components (**Fig. 4h**).

For the electrical installation section, LabVIEW (**Fig. 4h**), air compressor (**Fig. 4a**), Arduino (**Fig. 4g**), and DC power supply (**Fig. 4f**) can be connected to the 110V 50 Hz AC power supply through the corresponding live and neutral wires. Here, we leave the grounding segment of the 24V DC power supply empty, and connect the positive and negative terminals of the 24V supply to the No.9 and No.18 (COM) ports of the air filter regulator, respectively. Since No.18 is the common negative terminal, it also needs to be connected to the GND of the Arduino (**Fig. 4g**). We connect input signals No.1 to No.5 to the digital pins 1 to 5 of the Arduino. This setup allows us to control the output pressure of various ports of the air filter regulator through the Arduino. Considering the pressure sensors, we also need 24V power supply. Here, the brown and blue wires (**Fig. 4e**) of the pressure sensor are connected to the positive and negative terminals of the DC power supply, while the black wire serves as the output voltage signal, fed back to the analog input ports 1 to 5 of LabVIEW (**Fig. 4h**). Consequently, the sensor, DC power supply, and NI9205 GND need to be common ports. Additionally, the 24V DC power supply needs to be connected to the transistor array since our release valves are controlled by current, and this slightly higher current is managed by the transistor array. Therefore, we connect digital pins 6 to 10 of the Arduino to the I1 to I5 input ports of the transistor array (**Fig. 4m**). The O1 to O5 output ports of the transistor array can be directly connected to the input ports of the release valves. The negative terminals of the power supply and the signal terminals are shared. To simplify such complex wiring, we use a printed circuit board to interconnect all the negative terminals and handle the common 24V positive terminal connections.

Regarding the pneumatic connections, we first place the air compressor (**Fig. 4a**) and then connect it using pipes of specific sizes (**Fig. 4k**). Next, we install appropriate fittings in the piping system to facilitate proper shut-off processes and connect these to the filter regulator (**Fig. 4b**). Afterward, the piping is connected to the inlet of the electro-pneumatic regulator, leaving the outlet open. We connect the five output control ports of the selected electro-pneumatic regulator to the five input control ports of the release valves, which are subsequently connected to the pressure sensors. This allows the output ports of the pressure sensors to connect to our five actuators. Throughout this process, we use pneumatic adaptors (**Fig. 4l**) to assist in connecting the tubes, ensuring secure and reliable connections. After completing the installation of the basic components, we vertically install 150 mm short aluminum frames. Next, we install four 450 mm long aluminum frames to complete the basic framework of our stage. We fasten the perforated board onto the installed framework using fasteners. The stand, flush, and other components are secured onto the perforated board using bolt fasteners. Finally, we mount the rectal models that we need to test onto our stand and connect each actuator to them.

## 6. Operation instructions

After installing all the systems, we first connect the necessary power system to the external AC power supply. This allows components such as the air compressor, cDAQ-9178, Arduino, and pressure sensors to be in working condition. We install LABVIEW software on the computer and open the pre-designed program as shown in **Fig. 6**. After selecting the data storage location, sample rate, terminal configuration, and mapping the pressure input channels to their corresponding sensor ports, we can execute the start button. The data will then be displayed on the oscilloscope on the right side and saved to the computer. Next, we use the Arduino IDE software to upload the control code. Once written, we can control the regulator and release valve accordingly. Finally, we turn on the air compressor to 700 kPa. Under these conditions, the entire system can drive the five actuators.

## 7. Validation and characterization

We tested the pressure magnitude generated by the system. First, we wrote control code in the Arduino, which can control the output of the cDAQ-9178. Feedback testing was then conducted using pressure sensors **Fig. 7** illustrates the behavior of five parameters (P1 to P5) over a time range of 0 to 400s. Parameters P1, P2, and P3 exhibit minimal fluctuations around zero, indicating low variability. In contrast, P4 and P5 show significant oscillatory patterns with amplitudes around 30 kPa. This clear distinction implies different underlying dynamics or measurement scales for the two groups of parameters.

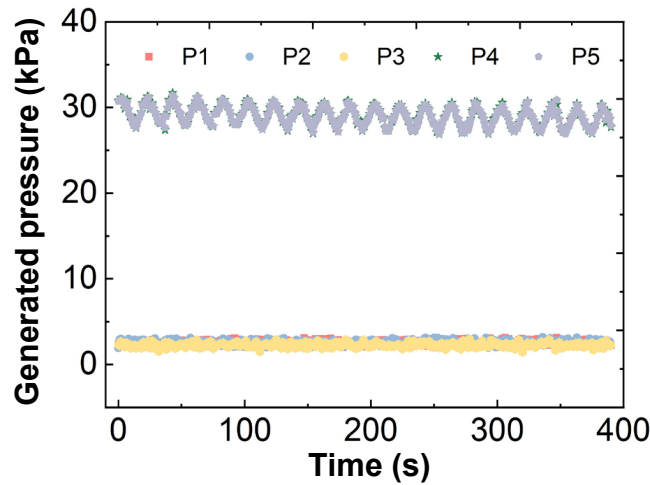

**Fig. 7** Pressure values obtained by DAQ system relative to five sensors

To validate the performance of simulating a sealed liquid environment in the simulated anus, akin to situations where loose stool does not automatically descend, we employed an actuator that occludes the anal opening to replicate human defecation (**Fig. 8 a & b**). It can be observed that our anal opening actuator demonstrates excellent sealing characteristics. Furthermore, our defecator can effectively simulate the division of feces into two parts, emulating functionality similar to that of the human anus (**Fig. 8 c& d**).

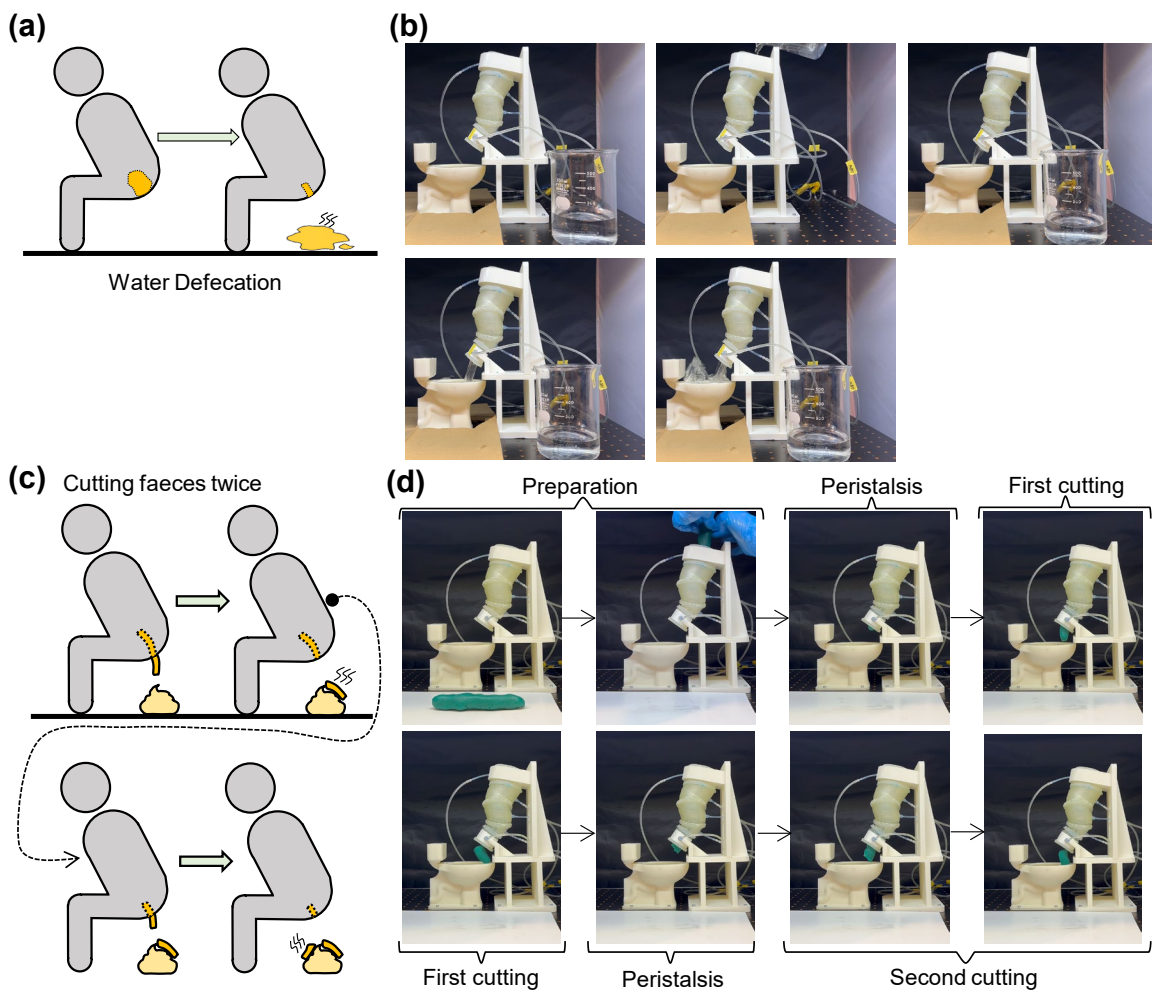

**Fig. 8 (a):** Functionality simulating human diarrhea; **(b):** Achieving the function of simulating human diarrhea by controlling the actuator of the anal opening; **(c):** Cutting feces twice through the anal opening. **(d):** Simulating the cutting of feces similar to the human body through peristalsis and the opening and closing of the anal opening.

## Ethics statements

We confirmed that our work does not involve any animal or human experiments.

## CRediT author statement

**Zebing Mao:** Conceptualization, Writing- Original draft, Software; **Sota Suzuki**: Data curation, Methodology, preparation. **Ardi Wiranata:** Software, Validation. **Ohgi Junji:** Reviewing and Editing. **Shoko Miyagawa**: Conceptualization, Supervision, Investigation.

## Acknowledgments

This work was supported by JSPS KAKEN (23K13290) and Ube City Next Generation Researchers, Japan.

# References

[1] Payne, C. J., Wamala, I., Bautista-Salinas, D., Saeed, M., Van Story, D., Thalhofer, T., Walsh, C. J. (2017). Soft robotic ventricular assist device with septal bracing for therapy of heart failure. Science Robotics, 2(12), eaan6736.

[2] D. Zrinscak, L. Lorenzon, M. Maselli, M. Cianchetti, Soft robotics for physical simulators, artificial organs and implantable assistive devices, Progress in Biomedical Engineering, 5(2023) 012002.

[3] Jiao, Z., Hu, Z., Shi, Y., Xu, K., Lin, F., Zhu, P., ... & Zou, J. (2024). Reprogrammable, intelligent soft origami LEGO coupling actuation, computation, and sensing. The Innovation, 5(1).

[4] Kong, D., Tanimura, R., Wang, F., Zhang, K., Kurosawa, M. K., & Aoyagi, M. (2024). Swimmer with submerged $SiO_2/Al/LiNbO_3$ surface acoustic wave propulsion system. Biomimetic Intelligence and Robotics, 4(2), 100159.

[5] Peng, Yanhong, et al. "Peristaltic Transporting Device Inspired by Large Intestine Structure." Sensors and Actuators A: Physical (2023): 114840.

[6] Zhao, L., Wu, Y., Yan, W., Zhan, W., Huang, X., Booth, J., ... & Balkcom, D. (2023). Starblocks: Soft actuated self-connecting blocks for building deformable lattice structures. IEEE Robotics and Automation Letters, 8(8), 4521-4528.

[7] E. T. Roche et al., "Soft robotic sleeve supports heart function," Science translational medicine, vol. 9, no. 373, p. eaaf3925, 2017.

[8] Hashem, R., Kazemi, S., Stommel, M., Cheng, L. K., & Xu, W. (2023). SoRSS: A Soft Robot for Bio-Mimicking Stomach Anatomy and Motility. Soft robotics, 10(3), 504-516.

[9] Fattorini, E., Brusa, T., Gingert, C., Hieber, S. E., Leung, V., Osmani, B., Müller, B. (2016). Artificial muscle devices: innovations and prospects for fecal incontinence treatment. Annals of biomedical engineering, 44, 1355-1369.

[10] Han, D., Yan, G., Wang, Z., Jiang, P., Liu, D., Zhao, K., & Ma, J. (2021). An artificial anal sphincter based on a novel clamping mechanism: Design, analysis, and testing. Artificial Organs, 45(8), E293-E303.

[11] van der Wilt, A. A., Breukink, S. O., Sturkenboom, R., Stassen, L. P., Baeten, C. G., & Melenhorst, J. (2020). The artificial bowel sphincter in the treatment of fecal incontinence, long-term complications. Diseases of the Colon & Rectum, 63(8), 1134-1141.

[12] Lehur, P.-A., McNevin, S., Buntzen, S., Mellgren, A. F., Laurberg, S., & Madoff, R. D. (2010). Magnetic anal sphincter augmentation for the treatment of fecal incontinence: a preliminary report from a feasibility study. Diseases of the Colon & Rectum, 53(12), 1604-1610.

[13] Liu, H., Luo, Y., Higa, M., Zhang, X., Saijo, Y., Shiraishi, Y., . Yambe, T. (2007). Biochemical evaluation of an artificial anal sphincter made from shape memory alloys. Journal of Artificial Organs, 10(4), 223-227.

[14] Shen, Y., Chen, M., & Skelton, R. E. (2023). Markov data-based reference tracking control to tensegrity morphing airfoils. Engineering Structures, 291, 116430.

[15] W.E. Stokes, D.G. Jayne, A. Alazmani, P.R. Culmer, A biomechanical model of the human defecatory system to investigate mechanisms of continence, Proceedings of the Institution of Mechanical Engineers, Part H: Journal of Engineering in Medicine, 233(2019) 114-26.

[16] Mao, Z., Iizuka, T., & Maeda, S. (2021). Bidirectional electrohydrodynamic pump with high symmetrical performance and its application to a tube actuator. Sensors and Actuators A: Physical, 113168.

[17] M. Cianchetti, C. Laschi, A. Menciassi, P. Dario, Biomedical applications of soft robotics, Nature Reviews Materials, 3(2018) 143-53.

[18] Tokoro, K., Hashimoto, T., & Kobayashi, H. (2014). Development of Robotic Defecation Simulator. Journal of Robotics and Mechatronics, 26(3), 377-387.

[19] Mao, Z., Suzuki, S., Nabae, H., Miyagawa, S., Suzumori, K., & Maeda, S. (2024). Machine-Learning-Enhanced Soft Robotic System Inspired by Rectal Functions for Investigating Fecal incontinence. arXiv preprint arXiv:2404.10999